\documentclass[11pt]{article}

\usepackage[preprint]{acl}

\usepackage{times}
\usepackage{latexsym}
\usepackage[T1]{fontenc}
\usepackage[utf8]{inputenc}
\usepackage{microtype}
\usepackage{inconsolata}
\usepackage{graphicx}
\usepackage{placeins}
\usepackage{flushend}
\usepackage{url}
\usepackage{booktabs}
\usepackage{tabularx}
\usepackage{array}
\usepackage{amssymb}

\newcolumntype{Y}{>{\raggedright\arraybackslash}X}

\usepackage{enumitem}
\setlist{itemsep=2pt,parsep=0pt,topsep=3pt}

\title{Interactive Training 2: Auditable Control Plane for Live Model Training}

\author{Wentao Zhang\textsuperscript{1,*} \quad
  Xuanhe Pan\textsuperscript{2,*} \quad
  Han Zhou\textsuperscript{1,*} \\
  \textbf{Yang Lu}\textsuperscript{2,\textdagger} \quad
  \textbf{Yuntian Deng}\textsuperscript{1,\textdagger} \\
  \textsuperscript{1}University of Waterloo \qquad
  \textsuperscript{2}University of Wisconsin-Madison \\
  \textsuperscript{*}Equal contribution \qquad
  \textsuperscript{\textdagger}Equal advising \\
  \texttt{\{w564zhan,h46zhou,yuntian\}@uwaterloo.ca} \quad
  \texttt{\{xpan73,ylu97\}@wisc.edu}}

\begin{document}
\maketitle

\begin{abstract}
Experiment trackers show how training is progressing, but changing a live run still usually requires trainer-specific code. We present \textbf{Interactive Training 2}, an open-source control plane for steering training through a shared protocol. Training applications declare which settings and actions they expose, humans and automated controllers submit requests through the same interface, and the training loop validates and applies them at safe control points. A customized Aim workspace combines live metrics and controls with a chronological record of requests and outcomes. We demonstrate the system across five NLP and reinforcement-learning workflows. The released code and traces provide a reusable foundation for auditable human- and agent-guided training.
\end{abstract}

\section{Introduction}
Researchers rarely treat model training as a completely passive process. They watch the loss curves, inspect evaluations, and decide whether to lower the learning rate, rebalance the data, save a checkpoint, or stop an unstable run. Modern experiment trackers make these decisions easier by showing what is happening, but they offer no general way to carry them out. Acting on an observation still usually requires custom callbacks or logic written for one trainer and one experiment. In practice, researchers already steer training in response to live behavior \citep{zhang2022optopenpretrainedtransformer,walsh2025}, but the tooling for doing so remains fragmented.

Interactive Training 2 makes a training run steerable through a shared interface. The training application declares which settings may be changed and which actions may be requested. A human, script, heuristic, or LLM agent can then submit a request through the same protocol. The training loop remains in control: it decides when the request can be safely applied, validates it, and records the result before continuing. For example, an application may expose a typed \texttt{learning\_rate} setting together with \texttt{evaluate} and \texttt{save\_checkpoint} actions. A controller may request a learning rate of $2\times10^{-5}$, but the optimizer is changed only when the loop reaches a designated control point. We call this layer between training code and its controllers a \emph{control plane}. It defines what can change, when changes take effect, and what is recorded.



Interactive Training v1 showed that a human or constrained LLM could modify a live Hugging Face training job through a control server \citep{zhang-etal-2025-interactive}. That system was built around a fixed vocabulary of commands executed through Hugging Face Trainer callbacks. Interactive Training 2 turns this trainer-specific demonstration into a reusable protocol. Training applications can register their own settings and actions, different training loops can choose their own control points, and different kinds of controllers can interact with all of them through the same interface.



\begin{figure*}[!t]
    \centering
    \includegraphics[trim={0.5cm 0.2cm 0 0},clip,width=\linewidth]{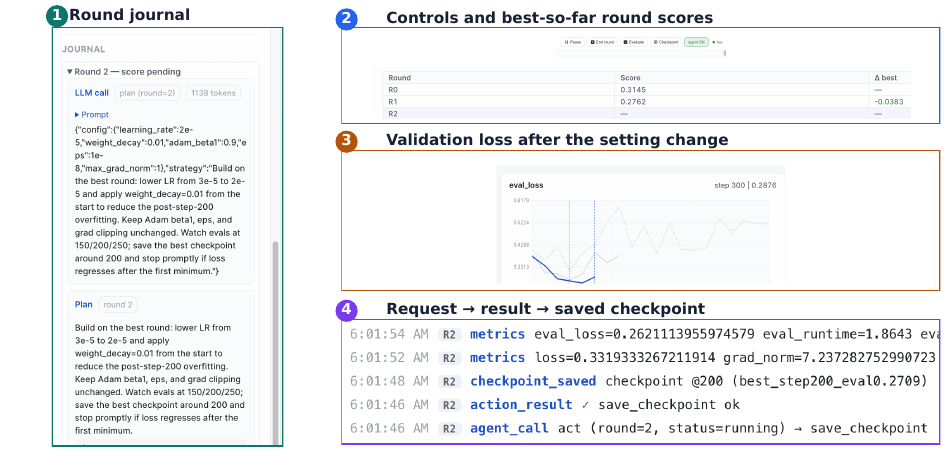}
    \caption{\textbf{One request from plan to result in the Aim Live workspace.}
    Enlarged crops from one illustrative BERT run show (1) the Round 2 journal and
    proposed change, (2) Controls and best-so-far round scores, (3) Validation loss
    after the setting change, and     (4) Logging area showing the request, result, and saved checkpoint.}
    \label{fig:interface}
\end{figure*}

Figure~\ref{fig:interface} previews this interaction in the customized Aim workspace. A researcher can inspect live metrics, request a change, and see whether and when it was applied. The accompanying journal connects plans and controller decisions to requests, results, evaluations, and checkpoints. We call the system \emph{auditable} because it is possible to reconstruct who requested each action, what the request contained, whether it succeeded, and what happened around it. The goal is to let researchers add human or automated steering to different training workflows without rebuilding the monitoring, control, and logging machinery for each experiment.

Our contributions are:
\begin{itemize}
\item \textbf{A shared protocol for steering live training runs.}
Training applications expose typed settings and structured actions, while training loops determine the safe points at which queued requests may take effect.

\item \textbf{A unified interface for observing, controlling, and reviewing training.}
A customized Aim workspace combines live metrics and controls with an ordered journal that connects controller decisions to requests, execution results, evaluations, and checkpoints.

\item \textbf{An open-source implementation demonstrated across five workflows.}
The same protocol supports callbacks, optimizer wrappers, and direct PyTorch and reinforcement-learning loops, and can be used by humans, scripts, and LLM agents. The code is available at \href{https://github.com/yuntian-group/interactive-training}{github.com/yuntian-group/interactive-training} and an interactive demo is available at \href{https://interactivetraining.ai/live}{interactivetraining.ai/live}.

\end{itemize}

\section{System at a Glance}
\label{sec:interface}

Consider a researcher fine-tuning BERT. From the Aim workspace, the researcher inspects the live validation trajectory, lowers the learning rate from $3\times10^{-5}$ to $2\times10^{-5}$, requests an evaluation and checkpoint, and lets training continue. The training loop applies these requests when it reaches a designated control point, records whether they succeeded, and then proceeds with the updated run.

Figure~\ref{fig:interface} illustrates this shared interaction pattern: observe the run, submit a request, apply it at a safe point, record the outcome, and continue. Humans, scripts, heuristics, and LLM agents all submit requests through the same protocol. The journal keeps each request together with its source, result, and surrounding training context, making the sequence of decisions and outcomes easier to inspect than when it is scattered across console logs.


\section{Control-Plane Design}
\label{sec:goals}

\paragraph{Training code defines each control.}
The application registers typed settings, which are called knobs in the Python API, and
implements the corresponding getters, setters, and actions.
The same mechanism can represent an AdamW \cite{loshchilov2018decoupled} learning rate, source-sampling weight,
Muon \cite{jordan2024muon} momentum, or curriculum difficulty in reinforcement learning without putting task-specific logic in
the training session.
HTTP clients, the UI, and automated controllers all see the same list.

\begin{figure*}[!t]
    \centering
    \includegraphics[width=\textwidth]{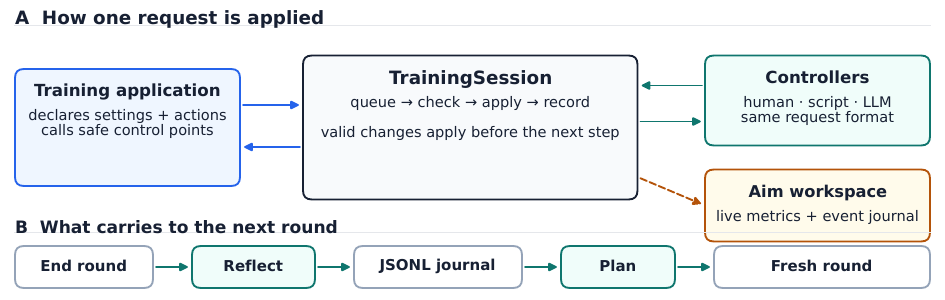}
    \caption{\textbf{The shared control protocol.}
    Training code declares settings and actions and calls \texttt{TrainingSession} at
    safe control points.
    Human users, scripts, and the optional LLM agent submit the same typed actions;
    the session validates, applies, and records them before the next training step,
    while Aim presents live metrics and the event journal.
    The lower panel shows what carries across rounds; each new round creates a fresh
    model and optimizer.}
    \label{fig:architecture}
\end{figure*}

Figure~\ref{fig:architecture} answers three questions: what the training code exposes,
who may submit requests, and when the session applies and records them.

\paragraph{The loop decides when changes apply.}
Requests may arrive at any time, but the training loop applies them one at a time at
explicit control points.
Human users, scripts, and LLM agents submit the same request format to one queue.
The application chooses the control points, validates task-specific actions, and can
hide dangerous actions from the LLM agent.

\paragraph{The journal links rounds.}
Within a round, metrics, requests, results, checkpoints, and controller decisions
share one ordered journal.
Across rounds, the demo scripts start a new model and optimizer, then pass a text
summary of configurations, actions, scores, and reflections into the next plan.


\section{How the System Works}
\label{sec:system}

Figure~\ref{fig:architecture} shows the training loop, the long-lived
\texttt{TrainingSession}, and the interfaces used by controllers.
The training code still owns every model update.
Human users, scripts, and the optional LLM agent send requests through the session.

\subsection{Registered Settings and Actions}

A typed \textbf{setting} binds a name to a getter and setter, with a data type,
optional range and step size, and a natural-language description.
The API method is named \texttt{register\_knob}. For numerical settings, the session
converts and clamps requested values before calling training-owned code.
An \textbf{action} defines a command such as evaluation or checkpointing.
Each submitted request records the action type, its arguments, its source, an
identifier, and a timestamp.
The HTTP API and LLM agent both receive the same list of registered actions.
Examples include \texttt{set\_knob}, \texttt{evaluate},
\texttt{save\_checkpoint}, \texttt{load\_checkpoint}, \texttt{pause}, and
\texttt{stop}.
Applications can add domain-specific actions, such as freezing a language model embedding component.

A \textbf{goal} names the metric used to score a round, such as the best validation
loss observed.
The current implementation either minimizes or maximizes that value.
An \textbf{event} records an increasing sequence number, event type, payload, round,
and time.
Metric reports, plans, LLM calls, setting changes, action results, checkpoints, and
reflections all enter the same journal as typed events.

\subsection{Applying Requests at Control Points}

At each \texttt{session.step(metrics)} call, the training process:
\begin{enumerate}
    \item Records new metrics and snapshots the initial setting configuration. 
    \item Runs attached controllers on their configured schedule, such as every 100
    steps. 
    \item Drains human and automated requests from one action queue. 
    \item Runs the matching handler, clamps setting values, and records a result for
    every request. 
    \item Continues to service resume and stop requests while paused.
\end{enumerate}
When a loop is idle, \texttt{session.pump()} processes pending requests without
recording a training metric.
The call returns a \texttt{StepControl} object that tells the integration whether to
stop, evaluate, save or load a checkpoint, reset a module, or update a setting.
The training integration translates these decisions into the local framework's
semantics.

Table~\ref{tab:settings} summarizes the five released workflows, how each one
integrates the session, what it exposes, and the scores recorded in its journal.
\begin{table*}[!t]
\centering
\footnotesize
\setlength{\tabcolsep}{3.5pt}
\begin{tabularx}{\textwidth}{@{}l Y Y Y Y@{}}
\toprule
\textbf{Workflow} & \textbf{Integration} &
\textbf{Settings and actions} & \textbf{Rounds, steps, call frequency} &
\textbf{Score (R0 $\rightarrow$ best)} \\
\midrule
Sentiment classification &
Hugging Face callback &
Schedule, AdamW, clipping, evaluation, checkpoint, stop &
11 rounds; 1,000 steps; eval/controller 50/100 &
Validation loss \(0.354 \rightarrow 0.220\) \\
\hline
Sentiment mixing &
Direct PyTorch loop &
Source weights, class weights, learning rate &
11 rounds; 2,000 steps; eval/controller 100/100 & Macro-F1 \(0.568 \rightarrow 0.656\) \\
\hline
Layerwise GPT &
Direct PyTorch loop &
27 embedding, block, head, and residual-group LRs &
11 rounds; 3,000 steps; eval/controller 100/300 &
Validation loss \(5.346 \rightarrow 5.055\) \\
\hline
Muon--AdamW GPT &
Direct PyTorch loop &
Both LRs, Muon momentum, decay, gradient diagnostics &
11 rounds; 3,000 steps; eval/controller 100/300 &
Validation loss \(4.797 \rightarrow 4.429\) \\
\hline
RLVR Countdown &
Direct RL loop &
Policy, clipping, and curriculum controls &
8 rounds; 100 iterations; eval/controller 10/10 &
Hardest accuracy \(0.032 \rightarrow 0.232\) \\
\bottomrule
\end{tabularx}
\caption{\textbf{What the five released workflows expose.}
Each row summarizes one multi-round run. Round~0 uses the original configuration without the
LLM agent, and every later round starts with a fresh model and optimizer.
Eval/controller frequencies are in steps (iterations for RLVR Countdown).
Scores give context for the associated journals.}
\label{tab:settings}
\end{table*}

The control point guarantees that when the call returns, every accepted change
is already in place before the next step.
An HTTP request does not wait for that point.
\texttt{POST /actions} returns an identifier immediately, and the later
\texttt{action\_result} event says whether the request was applied or rejected.
After reconnecting, a client can ask for recent missed events with a \texttt{since}
cursor.

\subsection{Three Ways to Connect Training Code}

\paragraph{Trainer.}
\texttt{make\_interactive(Trainer)} installs a callback that registers optimizer
controls, reports training and evaluation metrics, and maps session decisions to
Trainer control flags \citep{wolf2020transformers}.
When evaluation is scheduled, the callback waits to invoke the LLM agent until fresh
evaluation metrics are available.
This callback path underlies the BERT \cite{devlin-etal-2019-bert} sentiment classification demonstrations and requires only replacing the
Trainer class and passing a session.

\paragraph{Optimizer patching.}
For loops that are difficult to edit, an optional wrapper turns
\texttt{optimizer.step()} into a control point.
It saves the returned \texttt{StepControl} on the session for the loop to read.

\paragraph{Direct integration.}
Custom pretraining loops and reinforcement learning with verifiable rewards (RLVR)
loops call \texttt{session.step} directly.
The application chooses which diagnostics to report and how to implement checkpoint
or evaluation requests.

\subsection{Optional LLM Agent}

Any controller can use the shared protocol.
The supplied LLM agent demonstrates automated use in three phases.
\textbf{Plan} receives the task, goal, available settings, remaining rounds, and a
short summary of earlier configurations, scores, actions, and reflections.
It returns an initial configuration and strategy.
\textbf{Act} receives recent metric values, current settings, earlier decisions in the
round, prior reflections, and tools generated from the registered actions.
It may issue several tool calls or intentionally take no action.
\textbf{Reflect} receives the completed trajectory, action history, score, and earlier
lessons, then writes a short actionable lesson and a distinct next variation.

All recorded runs use GPT-5.5 with high reasoning effort as the agent.
When the agent is attached, the multi-round driver first runs one reference round
without it.
Each later round starts a fresh model and optimizer, creates a plan, runs training,
computes a score, and writes a reflection to the JSONL journal.
Each entry includes the round's starting configuration, best score and step,
successful actions, and reflection.
Prompts use at most the ten latest entries to bound context growth.

\begin{figure*}[!t]
    \centering
    \includegraphics[width=0.95\textwidth]{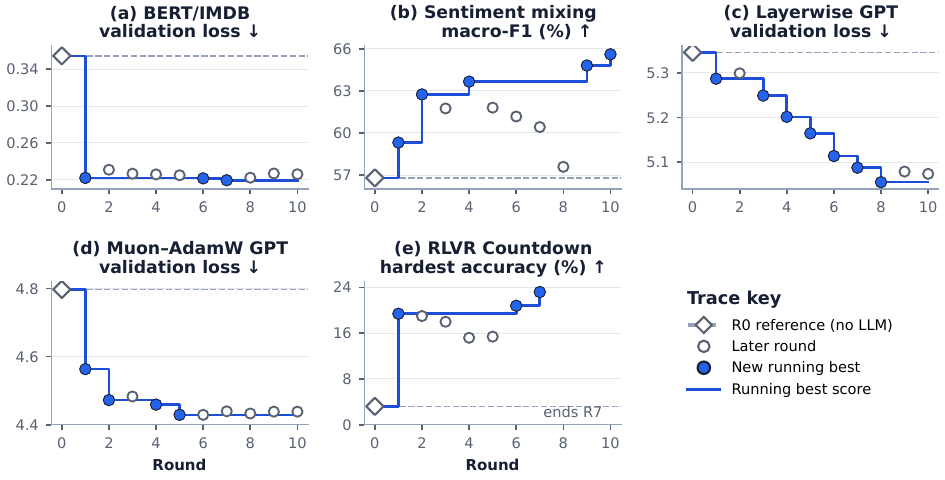}
    \caption{\textbf{Scores recorded during five example runs.}
    Each point is the best score recorded in one fresh round.
    The solid line tracks the best score seen so far, the diamond and dashed line mark
    the Round~0 reference without the LLM agent, and filled circles mark a new best.
    Y-axes use task-specific raw units and fixed limits. All panels share the
    Round~0--10 x-scale, with Countdown ending at Round~7.
    Sentiment and Countdown use percentages here. Table~\ref{tab:settings} reports
    their decimal values.}
    \label{fig:frontiers}
\end{figure*}

Figure~\ref{fig:frontiers} plots the score recorded for every fresh round.
Its purpose is to keep unsuccessful rounds visible alongside rounds that set a new
best.

Table~\ref{tab:memory-transitions} condenses three cases in which a failed or stalled
round changed the next plan.

\begin{table}[!t]
\begin{minipage}{\columnwidth}
\centering
\footnotesize
\begin{tabularx}{\columnwidth}{@{}l |Y@{}}
\toprule
\textbf{Workflow} & \textbf{Scores and next-plan evidence} \\
\midrule
Sentiment mixing & R8: 0.57586 $\rightarrow$ R9: 0.64812 $\rightarrow$ R10: 0.65612. Round 8 used a product-free \cite{10.1145/2507157.2507163}, tweet-dominant \cite{barbieri2020tweeteval} mixture and regressed. Its reflection proposed balanced tweet and finance \cite{Malo2014GoodDO} weights and asymmetric class weights. Round 9 records those choices explicitly. \\
\hline
Muon--AdamW GPT & R4: 4.4591 $\rightarrow$ R5: 4.4291. The Round 4 reflection names momentum as a setting it has not tried. The Round 5 plan lowers Muon momentum from 0.95 to 0.90. \\
\hline
RLVR Countdown & R5: 0.154 $\rightarrow$ R6: 0.208 $\rightarrow$ R7: 0.232. Round 5 starts at the hardest curriculum level and stalls. Its reflection proposes wider clipping and more early exploration; the Round 6 and 7 plans record those changes. \\
\bottomrule
\end{tabularx}
\captionsetup{hypcap=false}
\captionof{table}{Failed rounds appear in the next plan.}
\label{tab:memory-transitions}
\end{minipage}
\end{table}

\subsection{Aim and HTTP Interfaces}

A thin FastAPI service exposes \texttt{/state}, \texttt{/actions},
\texttt{/events}, and a WebSocket event stream.
It owns no training logic.
An Aim transport records events on a background thread so writing metrics does not
pause training.
It creates one Aim run per training round, stores the goal and control endpoint as run
metadata, records numerical metrics with round and branch context, and stores
controller-related events as text sequences.
The customized Aim workspace uses that endpoint both to display metrics and to submit
requests.

\section{System Validation}
\label{sec:evaluation}

\paragraph{Validation scope.}
We test three things.
First, can one protocol represent many kinds of changes?
Second, does each request produce a recorded result?
Third, can a later round visibly use an earlier reflection?
Scores provide context for the recorded decisions; they are not controlled
comparisons of optimization algorithms.
The artifact includes the five JSONL journals and the fields derived from
them.
Figure~\ref{fig:frontiers} provides the score context, and
Table~\ref{tab:settings} summarizes what each workflow exposes.


\subsection{Many Kinds of Changes}

The workflows range from optimizer and checkpoint settings to source and class
mixtures, 27 embedding, block, head, and residual-group learning rates, two optimizer
groups, and an RL difficulty schedule.
The same protocol works across these workflows: each application publishes its
settings and actions, and controllers read that list.
Appendix~\ref{app:tasks} and Table~\ref{tab:protocols} give model, data, budget, call
frequency, and diagnostic details.

\begin{table*}[!t]
\centering
\footnotesize
\setlength{\tabcolsep}{4pt}
\begin{tabularx}{\textwidth}{@{}l Y Y@{}}
\toprule
\textbf{Dimension} & \textbf{Interactive Training v1} & \textbf{This work} \\
\midrule
Training integration &
Hugging Face Trainer callbacks &
One \texttt{TrainingSession} used by Hugging Face callbacks, optimizer wrappers, and
direct loops \\
\hline
What can change &
Predetermined command types &
Application-registered typed settings and structured actions \\
\hline
When changes apply &
Execution at trainer callback boundaries &
Applied at developer-chosen control points before the next training step \\
\hline
Controllers &
Human dashboard and a limited learning-rate agent &
Humans, scripts, heuristics, and agents use one typed protocol \\
\hline
Interface &
Separate React dashboard &
Monitoring and control embedded in Aim \\
\hline
What is recorded &
Commands and checkpoint branches &
Ordered requests, results, metrics, checkpoints, plans, and reflections \\
\hline
What reaches the next round &
No explicit cross-round journal &
Explicit text journal across fresh rounds \\
\bottomrule
\end{tabularx}
\caption{\textbf{From a trainer-specific demonstration to a reusable control
protocol.}
Live intervention, an HTTP API, human control, checkpoint management, and
within-run LLM action are inherited from v1.}
\label{tab:v1-v2}
\end{table*}

Table~\ref{tab:v1-v2} compares this shared protocol with Interactive Training v1. 
Section~\ref{sec:related} returns to that distinction.

\subsection{From Request to Recorded Result}

Figure~\ref{fig:interface} exposes a complete request path in the deployed interface.
At 6:01:46, the LLM agent requests \texttt{save\_checkpoint}, the shared queue assigns
the request an identifier, the next control point runs the application handler,
and an \texttt{action\_result} records success.
Two seconds later, \texttt{checkpoint\_saved} records step 200 and evaluation loss
0.2709.
The same journal would record an explicit failure if validation or the handler
rejected the request.
Beyond this capture, the released JSONL journals record each round's executed
actions across all five workflows, so the request-to-result path can be checked
against the released traces.

\subsection{Using Earlier Reflections}

Table~\ref{tab:memory-transitions} gives three examples of a later plan using an
earlier reflection, while every round still starts a fresh model and optimizer.
After Sentiment R8 regresses, its reflection proposes asymmetric class weights;
Rounds 9 and 10 record those choices.
In Muon--AdamW, the R4 reflection says that momentum has not yet been tried.
The R5 plan lowers it from $0.95$ to $0.90$, preceding the score of $4.4291$.
After Countdown R5 begins at the hardest curriculum level and stalls at $0.154$, the
next plans widen clipping, increase early exploration, and sustain a mid-difficulty
curriculum before R6 and R7 reach $0.208$ and $0.232$.


\section{Related Work}
\label{sec:related}

Experiment trackers such as Aim and Weights \& Biases persist metrics
\citep{aim2023repo,biewald2020wandb}, while orchestration systems manage jobs and
resources.
Interactive Training 2 uses a tracker as its monitoring and control UI.
The training application, not the tracker, decides what may change.
Trainer callbacks can make similar changes inside one framework
\citep{wolf2020transformers}.
Our system instead gives every controller the same external protocol, records a
result for each request, and lets the training loop choose where changes apply.

Hyperparameter optimization, population-based training, and dynamic algorithm
configuration choose or adapt configurations according to a search policy
\citep{JMLR:v13:bergstra12a,jaderberg2017populationbasedtrainingneural,
akiba2019optuna,adriaensen2022dac,shabgahi2024livetunedynamicparametertuning}.
They can act as controllers on top of this system.
The control plane supplies the available actions, the places where changes apply,
and an audit trail; the search algorithm remains separate.

Recent LLM systems plan experiments or change training recipes
\citep{rao2026trainingmanager,radianis2026learnbywire,rodrigues2026llmhpo}.
In our system, the LLM agent is optional and uses the same protocol as humans,
scripts, and heuristics.
Interactive Training v1 is the direct predecessor; Table~\ref{tab:v1-v2} isolates the
new shared protocol from the inherited live-intervention demonstration.

\section{Conclusion}

Interactive Training 2 turns live training intervention from experiment-specific logic into a reusable interface. Training applications define what may change and when changes may safely take effect; humans and automated controllers decide what to request through a shared protocol, and the journal records the resulting sequence of decisions and outcomes. This separation makes training runs easier to steer, reuse, and inspect across different frameworks and workflows. The released system provides a foundation for interactive training tools and for more systematic studies of how humans and automated agents supervise learning in progress.

\section*{Limitations}


The training loop waits while the LLM responds. The system also does not yet handle a response that arrives after training has moved on. Applications remain responsible for checking custom actions. Value bounds, type checks, permissions, and human controls limit what an agent may do, but they cannot guarantee safe or optimal actions. High-stakes use needs approval gates, resource budgets, rollback policies, and task-specific safety checks.






\bibliography{custom}

\appendix
\clearpage

\section{Minimal Integration Example}
\label{app:integration}

The direct integration path requires a session, registered controls, and one call at a
safe point in the loop:

\begin{verbatim}
# 1. Create the training session.
session = TrainingSession(
    goal="val_loss",
    agent=LLMAgent(every=100),
    memory="memory.jsonl",
)
# 2. Register application-owned controls.
session.register_knob(
    "lr", get=get_lr, set=set_lr,
    min=0.0, max=1e-2,
)

with session.run():
    for step, batch in enumerate(loader):
        loss = update_model(batch)
        # 3. Apply at a safe boundary.
        control = session.step(
            {"loss": float(loss)},
            step=step,
        )
        if control.stop:
            break

# HF: make_interactive(Trainer)
\end{verbatim}

The artifact includes the complete \(3842\times1856\) unannotated Aim capture used
for Figure~\ref{fig:interface}.
It preserves the full journal, controls, metrics, and event-stream context at native
resolution; the public workspace is at
\url{https://interactivetraining.ai/live}.

\section{Control and Journal Records}
\label{app:records}

Every submitted request contains an action type, arguments, identifier, timestamp,
and source.
Every result records success or failure, optional returned data, and an error message.
Unknown actions and handler exceptions therefore appear as recorded failures.
Events add an increasing sequence number and round number, allowing reconnecting
clients to recover recent missed events; Aim keeps a permanent copy.
One cross-round journal row records whether the round is the Round~0 no-LLM
reference, its initial configuration and plan, best score and step, successful
actions, reflection, and cumulative token usage.
Prompts include at most ten recent rows.

\texttt{GET /state} returns the current settings, actions, and session metadata.
\texttt{POST /actions} enqueues a request whose later
\texttt{action\_result} reports success or failure; GET/WS
\texttt{/events?since=} returns events after a given sequence number.
Built-ins cover settings, evaluation, checkpoints, execution control, module reset,
annotations, context, and agent configuration.
Only humans may invoke destructive actions or actions that change the agent's own
configuration.

\section{Task Configurations and Controls}
\label{app:tasks}

The journals provide round-level plans, actions, reflections, scores, and usage.
Table~\ref{tab:protocols} lists the model, data, budget, evaluation frequency,
controller call frequency, and registered controls for each trace.
RLVR denotes reinforcement learning with verifiable rewards.

\begin{table*}[!t]
\centering
\footnotesize
\setlength{\tabcolsep}{3.5pt}
\begin{tabularx}{\textwidth}{@{}l Y l l Y@{}}
\toprule
\textbf{Workflow} & \textbf{Model and data} & \textbf{Budget} &
\textbf{Evaluate / controller (steps)} & \textbf{Settings, actions, and diagnostics} \\
\midrule
Sentiment classification &
BERT-base; 16k/1k IMDB \cite{maas-EtAl:2011:ACL-HLT2011} train/eval examples &
1,000 steps & 50/100 &
Schedule, AdamW LR/$\beta_1$/$\epsilon$, clipping, evaluation, checkpoint, stop \\
\hline
Sentiment mixing &
Tweet, finance, and product sources; 3k-example SST-5 \cite{socher-etal-2013-recursive} movie sentiment classification &
2,000 steps & 100/100 &
Three source weights, three class weights, LR \\
\hline
Layerwise GPT &
24$\times$512 Qwen3-style model; FineWeb-Edu \cite{lozhkov2024fineweb-edu} &
3,000 steps & 100/300 &
27 embedding, block, head, and residual-group LRs; layerwise gradient norms \\
\hline
Muon--AdamW GPT &
12$\times$768 Qwen3-style model; FineWeb-Edu &
3,000 steps & 100/300 &
Muon and AdamW LRs, momentum, decay; gradient norm and weight RMS \\
\hline
RLVR Countdown &
Qwen3-0.6B LoRA; $16\times8$ rollouts per iteration &
100 iterations & 10/10 &
LR, KL, temperature, clipping, entropy, difficulty \\

\bottomrule
\end{tabularx}
\caption{\textbf{Setup details for the five recorded runs.}
All recorded runs use GPT-5.5 with high reasoning effort as the LLM agent.
The public Muon video is a separate five-round run with 1,000 steps per round.
For RLVR Countdown, the evaluate and controller frequencies are in iterations.
The released journals do not include hardware, wall-clock time, or per-step
telemetry.}
\label{tab:protocols}
\end{table*}

\section{Released Evidence and Demonstration}
\label{app:demo}

\paragraph{Journal evidence.}
Table~\ref{tab:memory-ledger} reports execution and usage for each workflow.
The SHA-256 manifest links each row to its source journal.

\begin{table*}[t]
\centering
\scriptsize
\setlength{\tabcolsep}{4pt}
\begin{tabular}{@{}lrrrrrr@{}}
\toprule
\textbf{Workflow} & \textbf{Rounds} & \textbf{LLM rounds} &
\textbf{New bests} & \textbf{Actions} &
\textbf{Input/output tokens} & \textbf{Cost} \\
\midrule
BERT/IMDB & 11 & 10 & 3 & 88 & 251.6k / 86.1k & \$3.84 \\
Sentiment mixing & 11 & 10 & 5 & 399 & 905.5k / 185.7k & \$10.10 \\
Layerwise GPT & 11 & 10 & 7 & 355 & 1191.9k / 200.8k & \$11.98 \\
Muon--AdamW GPT & 11 & 10 & 4 & 137 & 449.8k / 138.4k & \$6.40 \\
RLVR Countdown & 8 & 7 & 3 & 228 & 429.3k / 69.0k & \$4.22 \\
\midrule
Total & 52 & 47 & 22 & 1207 & 3228.2k / 680.0k & \$36.54 \\
\bottomrule
\end{tabular}
\caption{\textbf{Per-workflow journal and usage summary.}
Token and cost totals come from the final row of each journal and are computed from
unrounded values. ``Actions'' counts successful setting changes and other control
actions, the count does not measure decision quality.}
\label{tab:memory-ledger}
\end{table*}

\paragraph{Trace shape.}
Figure~\ref{fig:frontiers} shows these patterns across all five workflows.
BERT makes its largest score change in the first LLM-guided round, whereas Layerwise
GPT continues setting new running bests through R8 and Muon--AdamW sets none after
R5.
Sentiment includes five later rounds that do not improve the current best before its
R9--R10 recovery; Countdown's R1 jump is followed by four such rounds before R6--R7.
The released plots retain these unsuccessful rounds rather than showing only the best
configuration.

\paragraph{Execution volume.}
Across the five sessions, 47 LLM-guided rounds record 1,207 summarized successful
actions, 3.23M/0.68M input/output tokens, and \$36.54 at the configured GPT-5.5
prices.
These fields quantify exposure and cost, not decision quality.

\paragraph{Live demonstration.}
The public site separates the five-round, 1,000-step-per-round Muon video
(\(5.44\!\rightarrow\!5.13\);
\url{https://youtu.be/SwEHORh6h0k}), the distinct seed-42 11-round,
3,000-step-per-round paper trace, and a queued tiny-BERT CPU sandbox driven by a
deterministic no-LLM policy.
In the sandbox, reviewers can change a setting or request an evaluation and observe
the matching \texttt{action\_result}, privately hosted LLM demos add the
plan/act/reflect path.

\section{Implementation and Reviewer Paths}
\label{app:reviewer-paths}

The optional LLM client supports OpenAI's Responses API and compatible chat
endpoints.
Provider, model, reasoning effort, and controller call frequency are runtime
settings; the recorded runs use GPT-5.5 with high reasoning effort.
API keys are write-only and are redacted from state and built-in action events.

\paragraph{Zero install.}
At \url{https://interactivetraining.ai/live}, reviewers can inspect a recorded trace or submit a limited set of setting changes and evaluation
requests to the queued CPU sandbox.
The separate Muon video is supplementary.

\paragraph{Core smoke test.}
Tag \texttt{v2.0.2} installs without the Aim fork; \texttt{tests/run\_tests.py}
exercises the session, action protocol, HTTP interfaces, built-in integrations,
journal, and scripted controllers without a GPU or provider key.

\paragraph{Full Aim workspace.}
To reproduce the full Aim interface, use the companion Aim branch and commit in
\texttt{demo/aim.lock.json}.
The main README gives the corresponding clone, \texttt{AIM\_SRC}, and BERT frontend
commands.

\end{document}